\documentclass[a4paper, 10pt, fleqn]{article}

\usepackage[a4paper, margin=1in]{geometry} 
\usepackage{graphicx} 
\usepackage{fancyhdr} 
\usepackage{titlesec} 
\usepackage{setspace} 
\usepackage{amsmath, amssymb} 
\usepackage{enumitem} 
\usepackage{hyperref} 
\usepackage{totpages}
\usepackage{caption} 
\usepackage{float}   
\usepackage{booktabs} 
\usepackage{algorithm} 
\usepackage{algpseudocode} 
\usepackage{IAAConference} 
\usepackage{subcaption}

\begin{document}
	
	\newcommand{\customauthor}[3]{\textbf{#1}\textsuperscript{#2}\ifnum#3=1\textsuperscript{*}\fi}
	
	\newcommand{\customaffiliation}[2]{\textsuperscript{#1}\textit{#2}}
	
	\newcommand{\authorslist}{

		\customauthor{Matthias Höfflin}{1}{1}, 
		\customauthor{Jürgen Wassner}{1}{0} 
	}
	
	\newcommand{\affiliationslist}{
		\customaffiliation{1}{Institute of Electrical Engineering, Lucerne University of Applied Science and Arts\\6048 Horw, Switzerland}}
	
	\begin{center}
		\textbf{}\\[1.5em]
		
		\textbf{\large Hardware-aware vs.\ Hardware-agnostic Energy Estimation for SNN\\ in Space Applications}\\[1.5em]
		
		\authorslist\\[0em]
		
		\affiliationslist\\[0em]
		
		\textit{* Corresponding Author}\\[2em]
	\end{center}

	\renewenvironment{abstract}{
		\begin{center}
			\textbf{Abstract}
		\end{center}
		\noindent\begin{spacing}{1}
		}{
		\end{spacing}\vspace{1em}
	}
	\begin{abstract}
Spiking Neural Networks (SNNs), inspired by biological intelligence, have long been considered inherently energy-efficient, making them attractive for resource-constrained domains such as space applications. However, recent comparative studies with conventional Artificial Neural Networks (ANNs) have begun to question this reputation, especially for digital implementations. This work investigates SNNs for multi-output regression, specifically 3-D satellite position estimation from monocular images, and compares hardware-aware and hardware-agnostic energy estimation methods. The proposed SNN, trained using the membrane potential of the Leaky Integrate-and-Fire (LIF) neuron in the final layer, achieves comparable Mean Squared Error ($MSE$) to a reference Convolutional Neural Network (CNN) on a photorealistic satellite dataset. Energy analysis shows that while hardware-agnostic methods predict a consistent $\sim$50–60 \% energy advantage for SNNs over CNNs, hardware-aware analysis reveals that significant energy savings are realized only on neuromorphic hardware and with high input sparsity. The influence of dark pixel ratio on energy consumption is quantified, emphasizing the impact of data characteristics and hardware assumptions. These findings highlight the need for transparent evaluation methods and explicit disclosure of underlying assumptions to ensure fair comparisons of neural network energy efficiency. 
	\end{abstract}
	
	\noindent \textbf{Keywords:} Spiking Neural Networks, Regression, Energy Estimation, Hardware-aware, Space Domain 

	\section{Introduction}
	Machine Learning (ML) and Artificial Intelligence (AI) have driven major advances in computer vision, with Convolutional Neural Networks (CNNs) excelling at tasks from digit recognition~\cite{ref1} to satellite pose estimation~\cite{ref2}. However, the high energy consumption of deep networks has prompted growing interest in brain-inspired alternatives, as the human brain is able to perform highly complex tasks with remarkable efficiency, operating on just 10–20 W of power — far less than most artificial systems. Spiking Neural Networks (SNNs) address this challenge by processing information as binary spikes, closely mimicking the brain’s energy-efficient mechanisms. Because SNNs dissipate less power with fewer spikes, they are particularly suitable for space imagery, which typically contains many dark pixels and therefore generates fewer spikes.
	
	\subsection{Problem statement}
	The main challenge in applying SNNs to regression tasks, such as position estimation of objects, lies in the binary nature of spikes, which contrasts with the continuous-valued activations and outputs typical of traditional Artificial Neural Networks (ANNs). Furthermore, although some studies claim that SNNs achieve energy efficiency by replacing Multiply-Accumulate (MAC) operations with simpler Accumulate (AC) operations, a closer look reveals significant challenges in realizing this theoretical advantage in practice. Therefore, establishing fair energy comparisons between ANNs and SNNs on the same task remains an active area of research.
	\subsection{Contributions}
	The main contributions of this work are as follows:
	\begin{itemize}
		\setlength{\itemsep}{0pt}
		\item We train a SNN for regression (position estimation of satellites) using a modified version of the approach introduced by Henkes et al.~\cite{ref4}, without employing a population layer.
		\item Based on Yan et al.~\cite{ref5}, an extended version of the energy estimation equation for Leaky Integrate-and-Fire (LIF) neurons in a neuromorphic dataflow architecture is presented.
		\item We provide a quantitative analysis comparing the strengths and weaknesses of our proposed energy estimation approach against that of Lunghi et al.~\cite{ref6}.
	\end{itemize}
	\section{Related work}
	\label{sec:related_work}
	\subsection{SNNs in regression tasks}
	While most SNN studies focus on classification, few address regression tasks. Henkes et al.~\cite{ref4} proposed a methodology for SNN-based regression in computational mechanics using various spiking neuron models, with decoding and population layers converting spike outputs into real values via membrane potentials. Gehrig et al.~\cite{ref7} used a global average spike pooling layer and non-spiking neurons to regress angular velocity from event-based input data.
	
	\subsection{Satellite pose estimation}
	Neural networks are frequently used for pose estimation of space objects~\cite{ref8}, but few studies have explored neuromorphic methods. Jawaid et al.~\cite{ref9} created the event-based SEENIC dataset but employed conventional neural networks for pose estimation. Courtois et al.~\cite{ref10} investigated  6-D pose estimation with SNNs on the SEENIC dataset, though their architecture included non-spiking, fully connected layers at the output. To date, a fully SNN-based approach to this problem has yet to be realized.
	
	\subsection{SNN energy estimation}
	The assumption that SNNs are inherently more energy-efficient than ANNs has been challenged by recent studies~\cite{ref5,ref11,ref12,ref13}, which suggest that the energy benefits of SNNs may be less significant than expected due to the overhead of multi-step processing and data movement. Comparative energy measurements can be biased by differences in implementation technologies, so analytical energy models aim to abstract from these factors using varying assumptions. Yan et al.~\cite{ref5} propose a hardware-aware methodology, while Lunghi et al.~\cite{ref6} present a hardware-agnostic approach. Although these studies go beyond simply counting synaptic operations by considering memory accesses and data dependencies, fair comparison of SNN and ANN energy consumption remains a challenge — as demonstrated by the results of this paper.
	 
	\section{Methodology}
	\label{sec:methodology}
	\subsection{SNN for position estimation}
	Equation~\eqref{eq1} describes the behavior of a LIF neuron from a deep learning perspective. Following~\cite{ref14}, $V_{l,i}[t]$ denotes the membrane potential of neuron $i$ in layer $l$ at time step $t$. This potential depends on the previous membrane potential $V_{l,i}[t-1]$, multiplied by a decay factor $\beta$, and the weighted input $X[t]$, where $X[t]$ is either $1$ or $0$. The equation also includes a reset mechanism that reinitializes the membrane potential after the neuron emits a spike, with $S_{l,i}[t]=1$ if $V_{l,i}[t]>V_{th}$, and $S_{l,i}[t]=0$ otherwise.
	\begin{equation}
		\label{eq1}
		V_{l,i}[t]= \beta V_{l,i}[t-1] + W X[t] - V_{th} S_{l,i}[t-1]
	\end{equation}
	In~\cite{ref4}, the authors proposed two specialized output layers: a penultimate decoding layer without a reset mechanism for continuous output, and a final population layer that computes the mean membrane potential from the decoding layer. In this work, we investigate 3-D satellite position estimation. Following the approach of~\cite{ref4}, the reset mechanism is removed from the final layer, and the output is taken from the last time step $t$ of the simulation window $T$, ensuring spike-based information propagation up to the last layer. To further enhance the network, the decay factor is set as a learnable parameter during training, following the neuron concept of~\cite{ref15}.
	\subsection{Energy estimation equations}
	Classical architectures (CAs) like CPUs and GPUs are ill-suited to the asynchronous binary spikes of SNNs, unlike their typical use for ANNs, making SNN energy estimation for such hardware inaccurate. To address this, the authors of~\cite{ref5} introduced an analytical equation for estimating the energy requirements of Integrate-and-Fire (IF) neurons in Neuromorphic Dataflow Architectures (NDAs) such as Loihi 2~\cite{ref16}. Equation~\eqref{eq2} extends their model to LIF neurons, where $N_{src}$ is the number of input connections, $T$ the simulation window, $s$ the sparsity rate, and $N_{hop}$ the number of routers a spike traverses in the architecture.
	\begin{equation} \label{eq2}
	\begin{aligned}
		E_{LIF} = &\ T \cdot\{\:\: N_{src} \cdot \underbrace{(1 - s_{in})}_{\text{firing rate}} \cdot \underbrace{\left(E_{R_{Weight}} + E_{ADD}\right)}_{\text{read weight \& add operations}} \\
		&+ \underbrace{E_{R_{state}}}_{\text{read state}} + \underbrace{E_{ADD} + E_{CMP}}_{\text{add \& compare}} + \underbrace{E_{R_{leak}}+E_{MUL}}_{\text{read \& multiply leakage factor}}+ \underbrace{(1 - s_{out}) \cdot E_{SUB}}_{\text{firing}} + \underbrace{E_{W_{state}}}_{\text{update potential}} \\
		&+ N_{src} \cdot (1 - s_{out}) \cdot \underbrace{N_{hop} \cdot E_{TPHop}}_{\text{data movement}}\:\:\}
	\end{aligned}
	\end{equation}
	\subsection{Experimental setup}
\begin{figure}[H]
		\centering
		\begin{subfigure}{0.22\textwidth}
			\includegraphics[width=\linewidth]{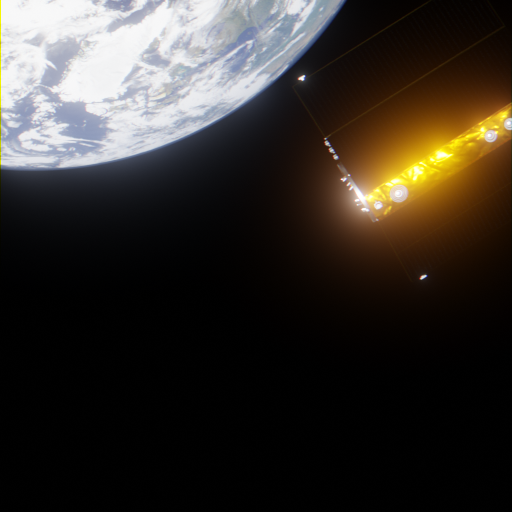}
		\end{subfigure}
		\begin{subfigure}{0.22\textwidth}
			\includegraphics[width=\linewidth]{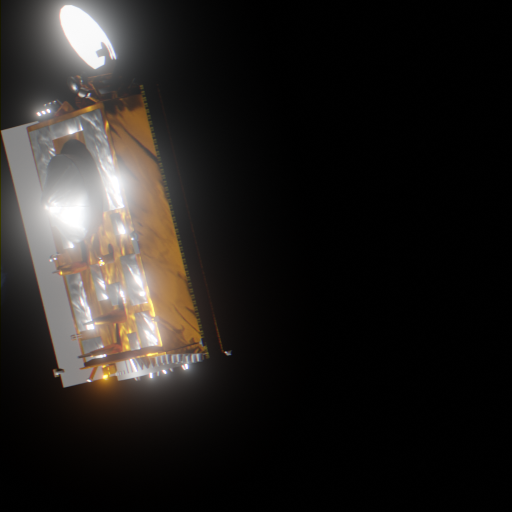}
		\end{subfigure}
		\begin{subfigure}{0.22\textwidth}
			\includegraphics[width=\linewidth]{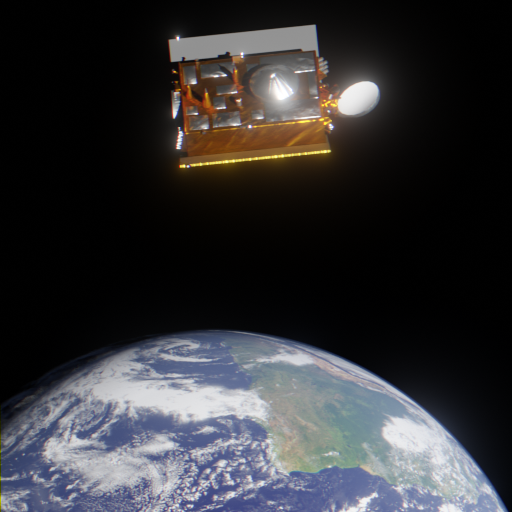}
		\end{subfigure}
		\begin{subfigure}{0.22\textwidth}
			\includegraphics[width=\linewidth]{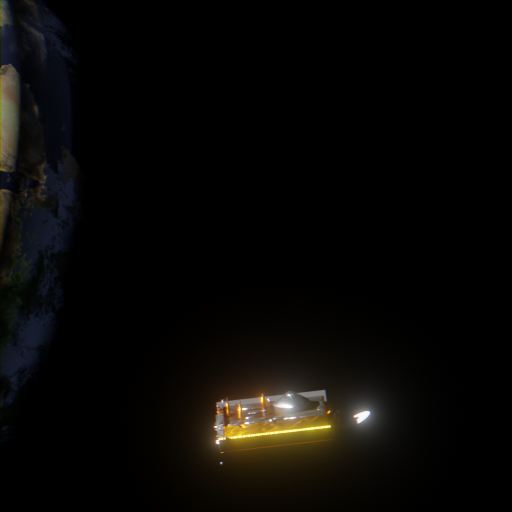}
		\end{subfigure}
		\caption{Example images from the test set}
		\label{fig:data}
	\end{figure}

	Collecting images from space is costly, making it challenging to build large datasets for neural network training. To address this, many deep learning applications use artificially generated images. In this work, a dataset of photorealistic images was generated, depicting the Sentinel-6 satellite~\cite{ref17}, Earth, and varying lighting conditions (see Fig.~\ref{fig:data}). Input images had a resolution of $256\times 256$ pixels with three color channels. The training and validation set contained 60,000 images representing 2,000 sequences of 30 images along different space trajectories, whereas the test set comprised 2,400 images from 20 sequences of 120 images each. For the SNN, input images were encoded as spike trains using direct encoding according to~\cite{ref18}, where the input image is fed to the first layer at every time step $t$ within the simulation window $T$. The SNN and reference CNN were trained using SpikingJelly~\cite{ref19} and PyTorch~\cite{ref20}, respectively. The architectures of SNN and CNN were identical, both for the convolutional head and dense layers. As this study focuses on SNNs for multi-output regression and energy efficiency, the 6-D pose estimation task was simplified to 3-D position estimation, predicting normalized $x, y, z$ coordinates of the satellite. Model performance was evaluated using Mean Squared Error ($MSE$).
	\subsection{Dark pixel ratio}
	One motivation for applying SNNs to satellite position estimation is that the dataset contains many dark pixels, which should result in fewer spikes and thus lower energy consumption. To quantify this, this work introduces the dark pixel ratio $\rho$, which relates the number of pixels $p$ in an image $\Omega$ with intensity $I(p)$ below a threshold $\theta$ to the total number of pixels, as shown in Eq.~\eqref{eq3}. The dark pixel ratio $\rho$ ranges from $0$ to $1$, representing the fraction of pixels with intensity below the threshold $\theta$. In this work, the intensity threshold $\theta$ was set to $0.05$ within the normalized intensity range $[0,1]$.
	\begin{equation}\label{eq3}
		\rho = \frac{|{p \in \Omega: I(p) < \theta}|}{|\Omega|}
	\end{equation}
	\section{Results}
	\label{sec:results}
	\subsection{Position estimation}
	Table~\ref{tab:results_mse} presents the $MSE$ of the trained SNN and CNN on the satellite position estimation task. To enable fair comparisons of energy consumption, both networks used identical architectures in terms of the number of layers and kernel sizes. While alternative architectures might achieve better performance, the primary goal here is to demonstrate that SNNs can effectively address regression tasks — such as satellite position estimation — with performance comparable to non-spiking CNNs.
	\begin{table}[h]
		\centering
		\caption{$MSE$ performance of CNNs and SNNs. (\textit{val} = validation dataset, \textit{test} = test dataset)}
		\label{tab:results_mse}
		\begin{tabular}{l|cc|cc}
			\toprule
			\textbf{Position Model} & \textbf{CNN (\textit{val})} & \textbf{CNN (\textit{test})} & \textbf{SNN (\textit{val})} & \textbf{SNN (\textit{test})} \\
			\midrule
			$MSE$ [$\times 10^{-3}$] & 51.2 & 55.2 & 54.5 & 62.0 \\
			\bottomrule
		\end{tabular}
	\end{table}
	\subsection{Energy estimation analysis}
	Table~\ref{tab:results_energy} shows energy estimates for relative comparison between hardware-agnostic and hardware-aware approach across different dark pixel ratios $\rho$ (col.\ 1). Col.\ 2 and 3 show the hardware-agnostic results in relative number of equivalent MAC (EMAC) according to \cite{ref6}. The remaining column show the results of our hardware-aware approach in units of relative energy using formulas from \cite{ref5} with our extension for LIF neuron in Eq.~\eqref{eq2}. The hardware-aware approach considers three different Memory External Ratios ($MER$), relating the energy demand of internal to external memory accesses for the Classical Architecture (CA), which is applicable to both CNN and SNN. The last column lists the estimated energy for the SNN on a Neuromorphic Dataflow Architecture (NDA), which uses only local memory accesses and thus is independent of $MER$.
\begin{table}[H]
	\centering
	\caption{Relative energy estimation results of SNN vs.\ CNN for dark pixel ratios $\rho$ of 10 different test sequences: hardware-agnostic (left) and hardware-aware (right) with different $MER$ and architectures.}			
	\label{tab:results_energy}
	\begin{tabular}{c|cc||cc|cc|cc|c}
		\toprule
		Test & \multicolumn{2}{c||}{\cite{ref6}} & \multicolumn{7}{c}{This work [rel.\ energy]} \\ \cmidrule{4-10}
		Seq. & \multicolumn{2}{c||}{ [rel.\ \# of EMAC]} & \multicolumn{2}{c|}{$MER$ 1:100 (\cite{ref21})} & \multicolumn{2}{c|}{$MER$ 1:50} & \multicolumn{2}{c|}{$MER$ 1:1} & - \\
		\cmidrule{1-3} \cmidrule{4-5} \cmidrule{6-7} \cmidrule{8-10}
		$\rho$& CNN & SNN & CNN\textsubscript{CA} & SNN\textsubscript{CA} & CNN\textsubscript{CA} & SNN\textsubscript{CA} & CNN\textsubscript{CA} & SNN\textsubscript{CA} & SNN\textsubscript{NDA}\\
		\midrule
		0.35 & 1.000 & 0.495 & 0.144 & 1.000 & 0.087 & 0.540 & 0.030 & 0.088 & 0.070 \\
		0.51 & 1.000 & 0.484 & 0.144 & 0.996 & 0.087 & 0.535 & 0.030 & 0.084 & 0.058 \\
		0.59 & 1.000 & 0.470 & 0.144 & 0.990 & 0.088 & 0.529 & 0.031 & 0.078 & 0.048 \\
		0.65 & 1.000 & 0.462 & 0.144 & 0.987 & 0.088 & 0.527 & 0.031 & 0.075 & 0.042 \\
		0.69 & 1.000 & 0.465 & 0.146 & 0.988 & 0.088 & 0.527 & 0.031 & 0.075 & 0.045 \\
		0.85 & 1.000 & 0.456 & 0.146 & 0.985 & 0.088 & 0.524 & 0.031 & 0.073 & 0.038 \\
		0.87 & 1.000 & 0.457 & 0.146 & 0.985 & 0.088 & 0.524 & 0.031 & 0.073 & 0.041 \\
		0.90 & 1.000 & 0.461 & 0.147 & 0.986 & 0.089 & 0.526 & 0.032 & 0.074 & 0.045 \\
		0.95 & 1.000 & 0.456 & 0.146 & 0.985 & 0.088 & 0.523 & 0.031 & 0.073 & 0.040 \\
		0.99 & 1.000 & 0.454 & 0.146 & 0.984 & 0.088 & 0.523 & 0.031 & 0.072 & 0.039 \\
		\bottomrule
	\end{tabular}
\end{table}%
	\noindent The results in Table~\ref{tab:results_energy} lead to the following conclusions:	
	\begin{itemize}
		\setlength{\itemsep}{0pt}
		\item The results in col.\ 2 and 3 confirm the findings of \cite{ref6}: SNNs possess an $\sim$50–60 \% efficiency advantage over CNNs when evaluated using a hardware-agnostic methodology. This methodology does not account for data dependencies in CNN computations, but it does reflect certain data dependencies present in the SNN model.
		\item The hardware-aware results in col.\ 4 to 9  indicate that SNNs do not surpass CNNs in terms of energy efficiency on classical computing architectures. As shown in col.\ 10, SNNs require neuromorphic hardware to achieve competitive energy efficiency compared to CNNs.				
		\item Whether SNNs can compete with CNNs in terms of energy efficiency depends strongly on the assumed energy ratio for different operations (Table~\ref{tab:results_energy} shows $MER$ as a primary example). When following~\cite{ref21} and using $MER=1:100$, SNNs show $\sim$50–80 \% lower energy demand than CNNs, which corresponds to the hardware-agnostic prediction. However, as $MER$ decreases, the energy advantage of SNNs diminishes and can ultimately reverse, making SNNs less efficient than CNNs.
		\item As expected, the energy demand of SNNs decreases with increasing dark pixel ratio, though not in a perfectly monotonic manner. This is because the spike rate, and therefore the energy demand, also depends on other factors, such as the edge density of image objects. While the dark pixel ratio primarily influences the convolutional head of the network, its direct impact on overall energy demand is clearly visible in Table~\ref{tab:results_energy}.
	\end{itemize}
	In summary, the hardware-agnostic method predicts a consistent $\sim$50 \% energy advantage for SNNs over CNNs. In contrast, the hardware-aware approach distinguishes between NDA and CA hardware, necessitating the selection of meaningful values for low-level mechanisms, such as the internal-to-external memory access ratio ($MER$), which may not always be readily available. However, with appropriate parameter selection, the hardware-aware method enables a deeper analysis of the factors influencing overall energy demand, thereby providing insight into energy optimization options.

	\section{Conclusions}
	\label{sec:conclusions}
	This work demonstrates the successful application of SNNs to multi-output regression for satellite position estimation, achieving comparable $MSE$ performance to CNNs with end-to-end spike processing. By eliminating the reset mechanism in the final layer and leveraging membrane potential, continuous regression outputs were obtained from the binary spiking network. The energy analysis compared hardware-agnostic and hardware-aware estimation methods. While the hardware-agnostic approach predicts a consistent $\sim$50 \% energy advantage for SNNs over CNNs with only moderate data dependency, the hardware-aware method reveals a stronger dependence on assumptions about low-level energy dissipation, such as memory accesses, and input spike rates. Consequently, hardware-aware estimation concludes that energy savings with SNNs depend on the use of neuromorphic hardware and increase with higher input sparsity.
	\section*{Acknowledgments}
	The research has been partially funded by the Swiss State Secretariat for Education, Research and Innovation, grant SBFI-633.4-2021-2024.
	\section*{AI disclosure}
	AI tools (DeepL/Perplexity) were employed for grammar correction and paraphrasing to improve manuscript clarity. All AI suggestions were reviewed and verified by the authors. AI was not used for data analysis, result interpretation, or generation of scientific content.
	

\end{document}